# Implementation of Fuzzy C-Means and Possibilistic C-Means Clustering Algorithms, Cluster Tendency Analysis and Cluster Validation

Md. Abu Bakr Siddique[1*], Rezoana Bente Arif[1#], Mohammad Mahmudur Rahman Khan[2@], and Zahidun Ashrafi[3$]
[1]Dept. of EEE, International University of Business Agriculture and Technology, Bangladesh
[2]Dept. of ECE, Mississippi State University, Mississippi State, MS 39762, USA
[3]Dept. of Statistics, Rajshahi College, Rajshahi 6100, Bangladesh
absiddique@iubat.edu[*], rezoana@iubat.edu[#], mrk303@msstate.edu[@], zimzahidun@gmail.com[$]

*Abstract*— **In this paper, several two-dimensional clustering scenarios are given. In those scenarios, soft partitioning clustering algorithms (Fuzzy C-means (FCM) and Possibilistic c-means (PCM)) are applied. Afterward, VAT is used to investigate the clustering tendency visually, and then in order of checking cluster validation, three types of indices (e.g., PC, DI, and DBI) were used. After observing the clustering algorithms, it was evident that each of them has its limitations; however, PCM is more robust to noise than FCM as in case of FCM a noise point has to be considered as a member of any of the cluster.**

*Keywords*— *Two-dimensional clustering, Soft clustering, Fuzzy c-means(FCM), Possibilistic c-means (PCM), cluster tendency, VAT algorithm, cluster validation, PC, DI, DBI, noise point.*

I. INTRODUCTION

The clustering [1-3] is a subfield of data mining technique and it is very effective to pick out useful information from dataset. Clustering technique is used to identify identical class of elements based on their characteristics [4]. Clustering is an unsupervised grouping technique which has a huge number of applications in many fields such as medicine, business, imaging [5], marketing, image segmentation [6], chemistry [8], robotics [7], climatology [9], etc. The best noted clustering techniques can be broadly classified into hierarchical [10-13], density based [16-19] clustering and partitioning clustering[12, 14, 15].

Unlike supervised classification where the labels of the learning data are provided, in case of unsupervised learning labels are not equipped with data set. The job of this paper is first to identify whether there is any cluster substructure in the data or not, then apply the clustering algorithm to separate the cluster and then, finally, VAT is used to observe the clustering tendency.

While approaching towards a clustering problem three issues must be taken care of:

1. We have to consider whether the data set has cluster substructure for $1 < c < n$. Here n is the number of data points in the dataset, and c is the cluster number.
2. After confirming the presence of substructure, we need to apply clustering algorithms make the computer identify the clusters. Several types of clustering algorithms can be used here, e.g., FCM, PCM, HCM and Mean shift.
3. After recognizing the clusters, cluster validity analysis should be applied to validate the clustering.

In this paper, all those issues were considered in step by step manner. In the case of FCM, a membership matrix was created which signifies the possibility of a data-point to be under a particular cluster. However, FCM has a constraint that for a specific data point the summation of the membership values for all the clusters must be '1' which makes FCM more vulnerable to noise. In contrast, this constraint was removed in case of PCM which made it more robust to noises.

II. CLUSTERING ALGORITHMS' BASICS

*A. The Basics of Fuzzy C-Means Algorithm*

In the Fuzzy c-means algorithm each cluster is represented by a parameter vector $\theta_j$ where j=1…c and c is the total number of clusters. In FCM, it is assumed that a data point from the dataset X does not exclusively belong to a single group; instead, it may belong to more than one cluster simultaneously up to a certain degree. The variable $u_{ij}$ symbolizes the degree of membership of $x_i$ in cluster $C_j$. The data point is more likely to be under the cluster for which the membership value is higher. The sum of all the membership value in all clusters of a particular data point must be 1. The algorithm involves an additional parameter q ($\geq 1$) which is called fuzzifier. The preferable value of the fuzzifier is 2. However, different other values were tried in this paper to observe the difference. The higher the value of the q, the less generalized the algorithm becomes.

FCM stems from the minimization of the cost function [20, 21]:

$$J(\theta, U) = \sum_{i=1}^{n}\sum_{j=1}^{c} u_{ij}^{q} \left\| x_i - \theta_j \right\|^2 \quad \ldots\ldots (1)$$

FCM is one of the most popular algorithms. It is iterative and starts with some initial estimates. Iteration contains following steps:

1. The grade of membership, $u_{ij}$ of the data $x_j$ in cluster $C_j$, i=1…N, and j=1…c, is computed taking into account the

Euclidean or Mahalanobis distance of $x_i$ from all $\theta_j$'s.

$$u_{ij} = \frac{1}{\sum_{k=1}^{c}\left(\frac{d(\vec{x}_i,\vec{\theta}_j)}{d(\vec{x}_i,\vec{\theta}_k)}\right)^{\frac{1}{q-1}}} \quad \ldots\ldots (2)$$

Then the representatives, $\theta_j$S are updated as the weighted means of all data vectors.

$$\theta_j = \frac{\sum_{i=1}^{n}(u_{ij})^q \vec{x}_i}{\sum_{i=1}^{n}(u_{ij})^q} \quad \ldots\ldots (3)$$

To terminate the algorithm, several methods can be applied. If the difference in the values of $\theta_j$S or the grade of membership between two successive iterations were small enough, the algorithm could be terminated. However, the number of iterations can be predetermined.

The FCM algorithm is sensitive in the presence of outliers because of the requirement of:

$$\sum_{j=1}^{m} u_{ij} = 1 \quad \ldots\ldots (4)$$

Equation (4) indicates even a noise point has to be considered to have a higher membership value in a particular cluster.

### B. The Basics of Possibilistic C-Means Algorithm

This algorithm (known as PCM) is also appropriate for unraveling compact clusters. It is a mode seeking algorithm. The framework here is similar to the one used in FCM. Each data vector $x_i$ is associated with a cluster $C_j$ via a scalar $u_{ij}$. However, the constraint that all $u_{ij}$s for a given $x_i$ sum up to 1 is removed. As a consequence, the $u_{ij}$s (for a given $x_i$) are not interrelated anymore, and they cannot be interpreted as a grade of membership of vector in cluster $C_j$ since this term implies that the summation of $u_{ij}$s for each $x_i$ should be constant. Instead, $u_{ij}$ is interpreted as the degree of compatibility between $x_i$ and $C_j$. The degree of compatibility between $x_i$ and $C_j$ is independent of that between $x_i$ and the remaining clusters.

PCM stems from the minimization of the cost function [22]:

$$J(\theta,U) = \sum_{i=1}^{n}\sum_{j=1}^{c} u_{ij}^q d(\vec{x}_i,\vec{\theta}_j) + \sum_{j=1}^{c} \eta_j \sum_{i=1}^{n}(1-u_{ij})^q \quad \ldots\ldots (5)$$

Here the first term of the equation is similar to the cost function of FCM. The second term was added because, without it, direct minimization for U leads to trivial zero solution and also this term ensure preferring large memberships.

As with FCM, a parameter q ($\geq$1) is involved in PCM. However, it does not act as a fuzzier in PCM. Like FCM, PCM is also iterative, and it even starts with some estimates.

1. The degree of compatibility, $u_{ij}$, of the data vector $x_i$ to cluster $C_j$, i = 1,...,N, j = 1,…,m, is computed, taking into account the (squared Euclidean) distance of $x_i$ from $\theta_j$ and the parameter, $\eta_j$.

2. The representatives, $\theta_j$S, are updated, as in FCM, as the weighted means of all data vectors ($u^q_{ij}$ weights each data vector $x_i$).

As like FCM, to terminate the algorithm, several methods can be applied. If the difference in the values of $\theta_j$S or the degree of compatibility between two successive iterations were small enough, the algorithm could be terminated. However, the number of iterations can be predetermined. In contrast to the FCM, PCM does not impose a clustering structure on input data X. PCM is sensitive to the initial $\theta_j$ values and the estimates of $\eta_j$ s.

In this paper, at first Fuzzy C-means algorithm was applied, which provided the cluster centers. Then the distance between each data point and the cluster center was measured. Now, if a data point exhibits minimum distance from cluster center k, then the data-point is considered to be under that cluster. By this theory, every data point was assigned to a particular cluster. While measuring the distance, both Euclidean and Mahalanobis distance were utilized.

### III. CLUSTER TENDENCY ANALYSIS

Cluster tendency analysis can be done by visually inspecting the reordered distance matrix of the given dataset known as the visual assessment of cluster tendency (VAT) [23] algorithm. In VAT algorithm, at first, the Euclidean distance matrix between the samples is computed. Then, this distance matrix is reordered to create an ordered dissimilarity matrix such that similar data points are located close to each other. Then this ordered dissimilarity matrix is converted to an ordered dissimilarity image, known as VAT image. In the picture, dissimilarity is represented by each pixel. If the image is scaled on the gray intensity scale, then, white pixels values show high contrast and black pixels exhibit low dissimilarity which is evident from the diagonal pixels where the entry of divergence is zero because dissimilarity is measured within the same data points. But VAT has a problem of being computationally expensive and limitation for discovering more sophisticated patterns inside the data. Therefore, improved VAT (iVAT) [24] can also be utilized. In the iVAT algorithm, a graph-theoretic distance transform has been used to enhance the effectiveness of the VAT algorithm for complex cases where VAT fails to provide detailed and clear cluster tendency.

### IV. CLUSTER VALIDATION

After cluster tendency analysis, cluster validity analysis was applied which ensures the validity of the number of clusters considered in the clustering algorithm. Cluster tendency provides a visual representation of the number of clusters; on the other hand, cluster validity analysis offers numerical value for different groups' validity indices which indicate the number of clusters. In this paper, three cluster validity indices were utilized for cluster validation.

1. Partition Coefficient (PC): It uses the membership matrix to compute the index as shown in equation (6) [25].

$$PC = \frac{1}{N}\sum_{i=1}^{C}\sum j = 1nu_{ij}^2 \ \ \ldots\ldots\ (6)$$

The higher value of PC indicates the more valid clustering number.

2. Dunn Index (DI): Dunn's cluster validity indices try to identify the compact set of points from small amount dispersion among the members of the same cluster and the different set of clusters by separating the distance measurements. The equation for Dunn Index can be written as equation (7) [26].

$$DI = min_{i \in C} min_{j \in C \& j \neq i} \frac{\delta(A_i, A_j)}{max_{i \in C} \Delta(A_k)} \ \ \ldots\ldots\ (7)$$

The Higher value of DI indicates better validation of the cluster.

3. Davies Bouldin Index (DBI): DBI considers the average case of each cluster by utilizing the mean error of each cluster. It introduces a scattering measure $S_i$ to measure the scattering within the same group. It maximizes the ratio of this scattering measure to the cluster center separation to give us the DBI for many clusters, C. Therefore, the equation for DBI can be expressed as equation (10) [27].

$$S_{i,q} = \left(\frac{1}{|A_i|}\sum_{x_j \in A_i} \|x_j - v_i\|_2^q\right)^{\frac{1}{q}} \ \ \ldots\ldots\ (8)$$

$$R_i = max_{j \in C \& j \neq i} \frac{S_{i,q} + S_{j,q}}{\|v_i - v_j\|} \ \ \ldots\ldots\ (9)$$

$$\therefore DBI = \frac{1}{c}\sum_{i=1}^{c} R_i \ \ \ldots\ldots\ (10)$$

From equation (10) it is visible that the value of DBI indicates the ratio of intra-class scattering to interclass separation. Hence, the lower value of DBI means better clustering.

V. RESULTS AND DISCUSSION

*A. The Performance of Fuzzy C-Means Clustering Algorithm*

After building the algorithm of Fuzzy C-means, different data sets were implemented (e.g., two separately spaced clusters, three close clusters, 4 clusters, 5 clusters, clusters with noise, etc.). Fuzzy C means algorithm performs reasonably well with clusters with no noise and separately spaced clusters.

Followings are some examples of the performance of the FCM clustering algorithm:

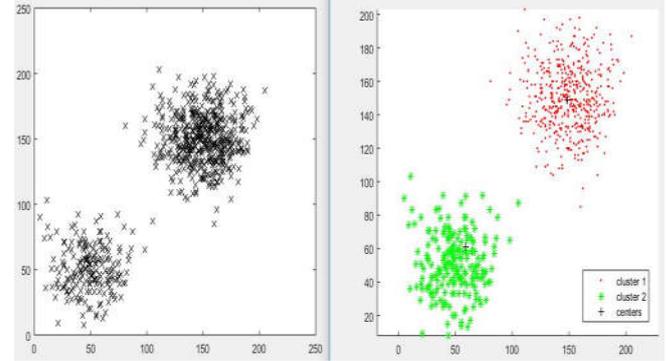
Figure 1: Two clusters and FCM output

Then differently dense data was implemented, and the FCM analysis was the following:

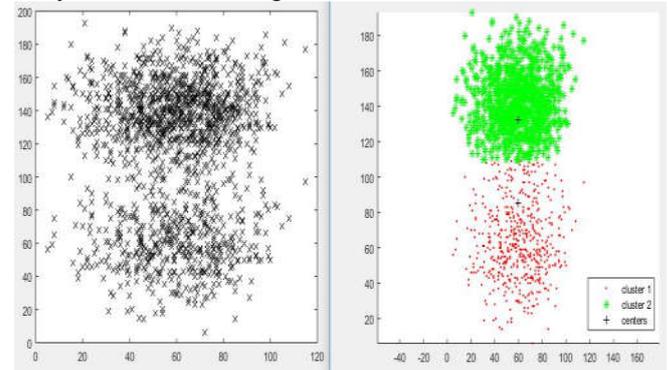
Figure 2: Two clusters and FCM output

Later on, a data set was implemented which has a different cluster size.

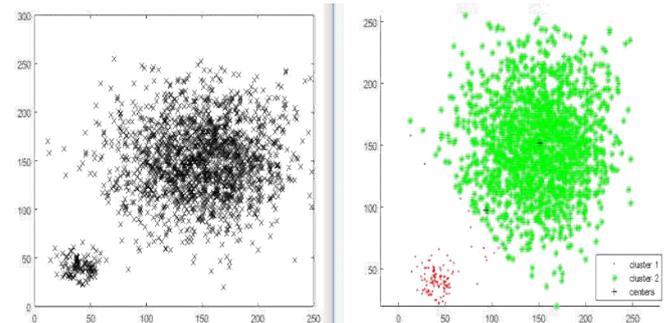
Figure 3: Two clusters and FCM output

In addition to that two close cluster data with different density and different size were implemented and the FCM showed the following result.

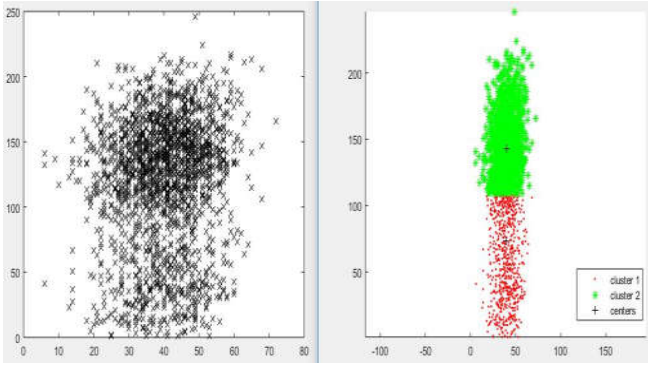
Figure 4: Two clusters and FCM output

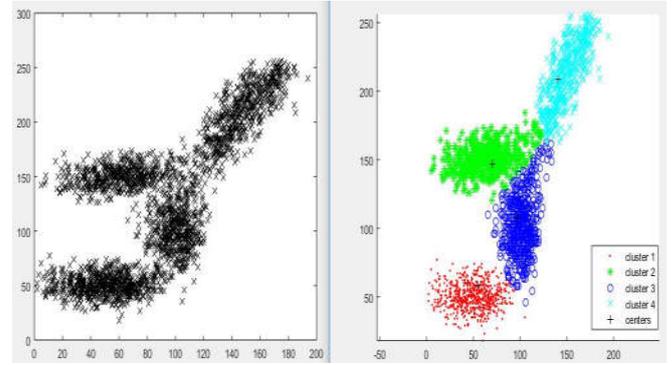
Figure 7: The dataset with four clusters and FCM output

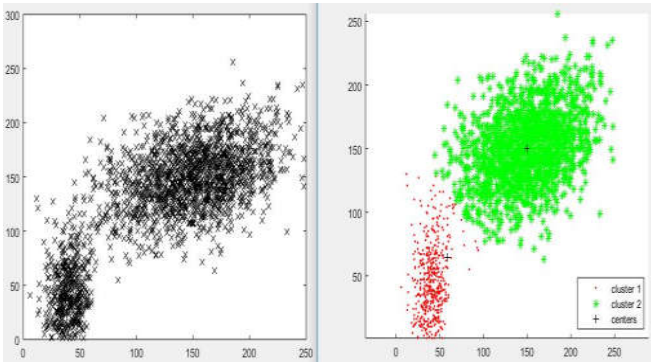
Figure 5: Two clusters and FCM output

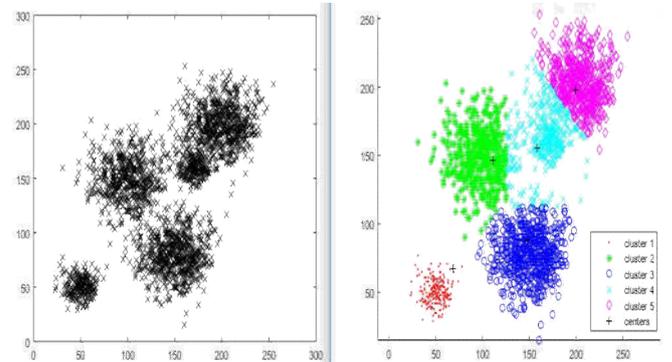
Figure 8: The dataset with five clusters and FCM output

As it is evident that FCM works well with data sets with two clusters, hence data sets with 3, 4 and 5 clusters were implemented afterward to observe the performance of FCM.

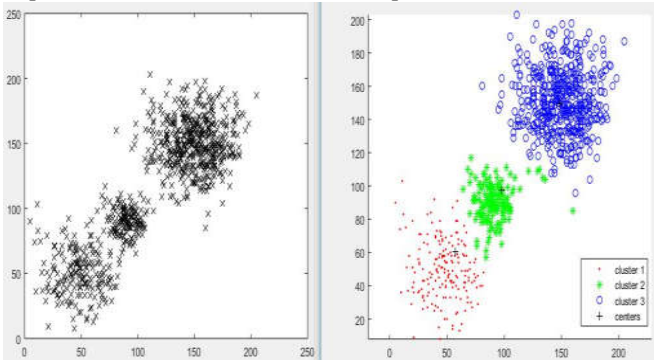
Figure 6: The dataset with three clusters and FCM output

*B. The Performance of Possibilistic C-Means Clustering Algorithm*

As like FCM, the same datasets were implemented in the PCM algorithm. At first, the data set with two well-separated clusters was performed, and the response from the PCM algorithm was following.

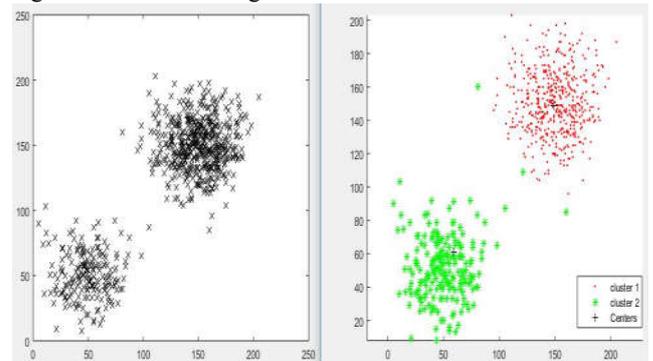
Figure 9: The dataset with two clusters and PCM output

PCM was able to find the two clusters. Then another dataset was implemented where there were two not wholly separable data.

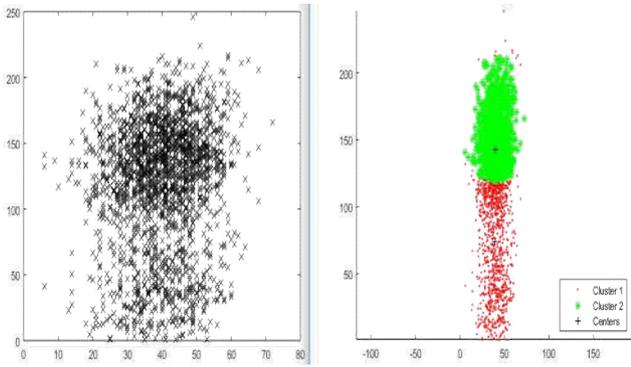

Figure 10: The dataset with two differently dense clusters and PCM output

It is evident that PCM was capable of clustering the data; however, it also miss-clustered some of the data. Next, data with three clusters was applied.

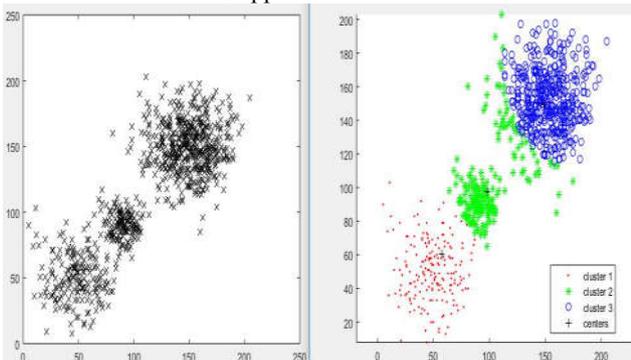

Figure 11: The dataset with three clusters and PCM output

After implementing three cluster data, four cluster data was used the PCM nicely clustered this.

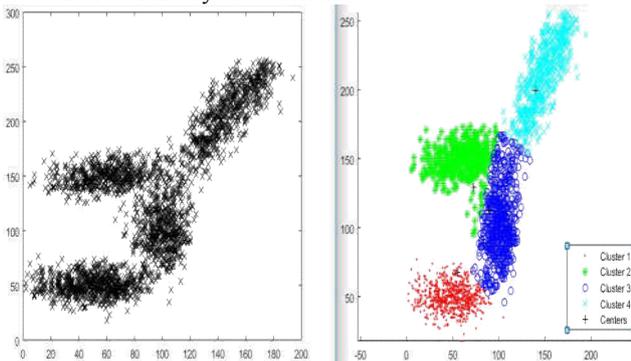

Figure 12: The dataset with four clusters and PCM output

Finally, the PCM algorithm was applied on a compact five cluster data and PCM performed reasonably well in clustering the data.

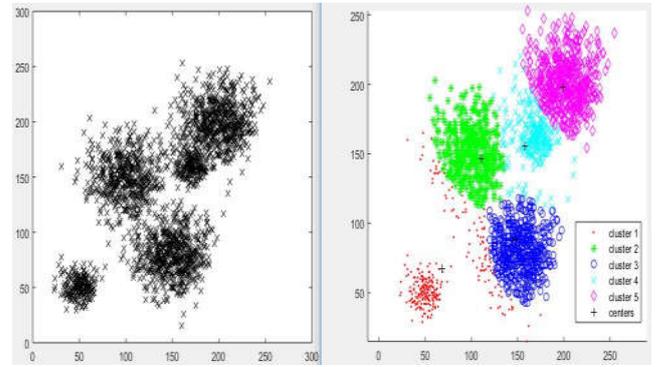

Figure 13: The dataset with five clusters and PCM output

## VI. Implementation of Cluster Validation Analysis

After implementing the data sets into different clustering algorithm, cluster validation analysis was applied. For this reason, three types of indices were utilized namely Partition Coefficient Index, Dunn's Index, and Davies-Bouldin's Index. Those indices showed the different result for different algorithms.

### A. The Performance of Fuzzy C-Means Algorithm

Table 1: PC, DI and DBI indices for two clusters FCM

| C | 2 | 3 | 4 | 5 | 6 |
|---|---|---|---|---|---|
| PC | 0.9193 | 0.6561 | 0.5276 | 0.4405 | 0.3537 |
| DI | 0.0453 | 1.2251e-04 | 7.1174e-05 | 7.1174e-05 | 3.5587e-04 |
| DBI | 0.2302 | 1.6054 | 3.5904 | 16.9824 | 8.3297 |

Table 2: PC, DI and DBI indices for three clusters FCM

| C | 2 | 3 | 4 | 5 | 6 |
|---|---|---|---|---|---|
| PC | 0.5005 | 0.7893 | 0.3421 | 0.2266 | 0.1935 |
| DI | 0.0032 | 0.0040 | 0.0012 | 7.6628e-04 | 2.3578e-04 |
| DBI | 24.8874 | 0.4901 | 1.9941 | 2.4549 | 8.1743 |

Table 3: PC, DI and DBI indices for four clusters FCM

| C | 2 | 3 | 4 | 5 | 6 | 7 |
|---|---|---|---|---|---|---|
| PC | 0.5032 | 0.3346 | 0.6651 | 0.2015 | 0.1673 | 0.1431 |
| DI | 2.5362e-05 | 2.6058e-05 | 0 | 2.6443e-05 | 2.6421e-05 | 2.6058e-05 |
| DBI | 14.4144 | 76.2557 | 0.4839 | 51.9591 | 303.9339 | 1.7369e+03 |

Table 4: PC, DI and DBI indices for five clusters FCM

| C | 2 | 3 | 4 | 5 | 6 | 7 |
|---|---|---|---|---|---|---|
| PC | 0.4489 | 0.6333 | 0.2500 | 0.4733(.2) | 0.1667 | 0.1429 |
| DI | 0 | 0 | 4.8019e-05 | 4.8210e-04 | 0 | 0 |
| DBI | 2.2617 | 3.8552e+06 | 1.1599e+06 | 1.6343 | 6.6217e+05 | 7.6173e+05 |

From the tables above, it is visible that for well separated and less number of clusters the clustering index performs reasonably well. And the performance of DBI is more accurate than PC and DI.

B. *The Performance of Possibilistic C-Means Algorithm*

Table 5: PC, DI and DBI indices for two clusters PCM

| C | 2 | 3 | 4 | 5 |
|---|---|---|---|---|
| PC | 1.6604 | 0.9799 | 1.3024 | 1.6378 |
| DI | 1.2206e-04 | 2.1865e-05 | 2.1865e-05 | 6.1256e-05 |
| DBI | 0.3167 | 2.2313 | 3.2730 | 3.2142 |

Table 6: PC, DI and DBI indices for four clusters PCM

| C | 2 | 3 | 4 | 5 | 6 |
|---|---|---|---|---|---|
| PC | 0.5340 | 0.3539 | 0.4037 | 0.5853 | 0.5068 |
| DI | 3.4204e-05 | 2.6443e-05 | 8.8252e-05 | 4.6622e-05 | 7.0018e-05 |
| DBI | 401.8789 | 6.4079 | 0.8122 | 18.8765 | 10.3166 |

Table 7: PC, DI and DBI indices for five clusters PCM

| C | 2 | 3 | 4 | 5 | 6 | 7 |
|---|---|---|---|---|---|---|
| PC | 0.4518 | 0.6777 | 0.9037 | 1.1296 | 1.3555 | 1.5814 |
| DI | 1.1012e-05 | 1.1012e-05 | 2.2024e-05 | 5.5060e-05 | 2.2024e-04 | 1.1012e-05 |
| DBI | 3.4813e-03 | 1.9671e-03 | 1.7894e-03 | 1.7385e-03 | 1.0792e-03 | 1.8421e-03 |

The cluster validity indices showed a perfect result for less number of clusters in case of PCM. From the table, it is visible that both DI and DBI are showing better performance in those data sets.

VII. IMPLEMENTATION OF CLUSTER TENDENCY ANALYSIS

For cluster tendency analysis, VAT was used. The results of cluster tendency analysis for both Fuzzy c-means and possibilistic c-means clustering are shown in table 8.

From table 8, it is evident that if the data set has well-separated clusters, then cluster tendency analysis will show a neat result. With the decrease of separation and increase of cluster number, the tendency analysis shows an ambiguous result. For example, in the case of four clusters, cluster tendency analysis for PCM was not showing a good pattern.

Table 8: Cluster tendency analysis for FCM and PCM

| | FCM | PCM |
|---|---|---|
| Two Separate Clusters | 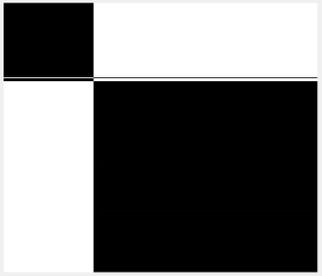 | 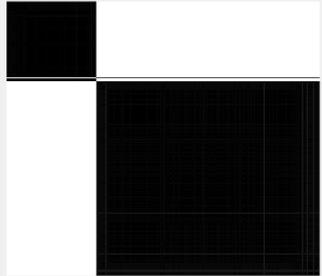 |
| Two different elliptic density | 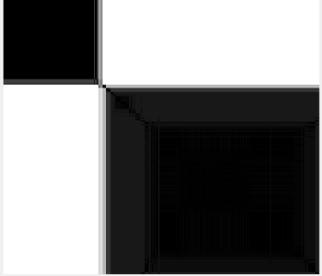 | 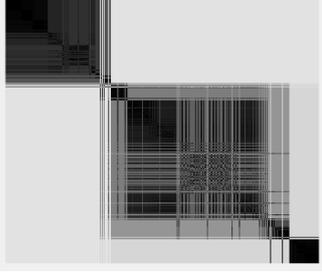 |
| Three close clusters | 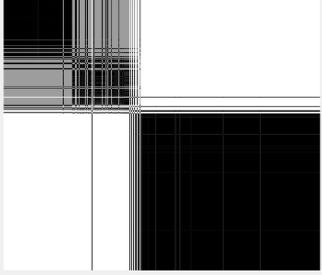 | 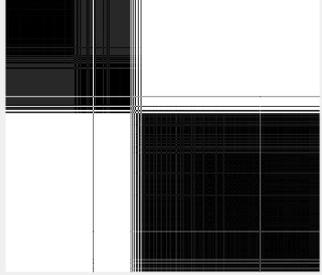 |
| Four clusters | 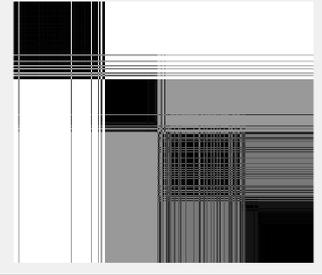 | 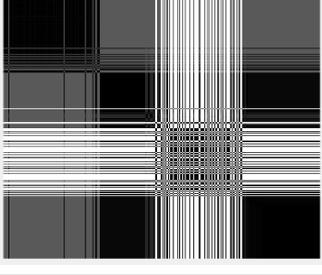 |
| Five clusters | 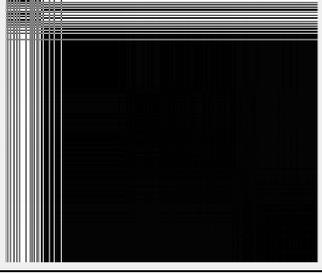 | 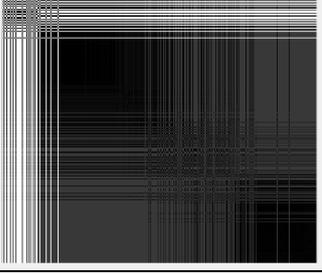 |

## VIII. CONCLUSION

This paper has discussed some aspects of cluster analysis process. From the analysis measures done in this paper indicates some significant observations. In the case of two separately places clusters, both FCM and PCM showed reasonably good performance. However, if noise is added to the dataset, then FCM may lead to a wrong result as the noise point shifts the cluster centers. On the other hand, PCM is free from this lacking as it does not consider the constraint of FCM. PCM is mode searching algorithm which makes it more stable in case of noisy data.